\documentclass[runningheads]{llncs}

\usepackage{orcidlink}
\usepackage{xurl}
\usepackage{multirow}

\usepackage[T1]{fontenc}
\usepackage{graphicx}
\usepackage{amsmath}
\usepackage{amssymb}
\usepackage{marvosym}

\begin{document}
\title{Size Matters: Foundation Model for Czech HTML documents}
%
%
\author{
Martin Dvořák \orcidlink{0009-0004-3845-7505} \thanks{The authors contributed equally and performed the majority of the work.} \and
Vít Tlustoš (\Letter) \orcidlink{0009-0009-3434-6441} $^{\star}$ \and
Artyom Voronin \orcidlink{0009-0008-7881-8941} $^{\star}$ \and
Martin Habrovec \orcidlink{0009-0004-5333-4714} \and
Kateřina Podlesná \orcidlink{0000-0002-1950-6744} \and
Barbora Rišová \orcidlink{0009-0005-0752-8806} \and
Josef Vonášek \orcidlink{0009-0006-9429-9278}
}
\authorrunning{Dvořák et al.}
%
\institute{Seznam.cz, Prague, Czech Republic\\
    \email{vit.tlustos@firma.seznam.cz}}

\maketitle              
\begin{abstract}
Creating universal, high-quality representations of web documents in high-traffic industrial environments requires models that are both performant and economic. Existing approaches, however, often depend on large models, overlook the structural information inherent in HTML, or are constrained by short context windows, limiting their ability to process real-world web pages. We present HTML-LM, a compact foundation model with 154 million parameters that addresses these limitations through HTML-aware training and a ModernBERT-based architecture. It was trained on 100 million web documents using multiple objectives, including masked language modeling, bag-of-words prediction, and contrastive distillation from large language models. Consequently, HTML-LM sets a new state-of-the-art for classification and regression applications in the Czech Internet domain, surpassing both larger encoders and small-sized LLMs. The model is deployed in production, processing thousands of web documents per second, and released to the community under the CC BY-NC 4.0\footnote{\url{https://creativecommons.org/licenses/by-nc/4.0/deed.en}} license.
\begin{center}
    \textbf{\url{https://huggingface.co/Seznam/html-lm}}
\end{center}

\end{abstract}
\section{Introduction}
In information retrieval systems, categorical metadata and document-level signals play a critical role in ensuring retrieval quality from the Internet. Signals such as document type~\cite{SearchingBasedOnRelevance}, spam likelihood~\cite{SpamToIR}, or the presence of adult content are essential for maintaining clean search indexes and improving search ranking and user satisfaction. However, at scale, incorporating these signals into real-time retrieval pipelines requires representations that are both computationally efficient and semantically expressive. An effective strategy is to encode each document into a compact embedding that serves as a shared representation for multiple downstream models.

Prior research into HTML document representation has evolved along several trajectories. Structure-aware models based on XPath-like encodings, such as MarkupLM~\cite{markuplm}, Structor~\cite{structor}, and DOM-LM~\cite{DOMLM}, treat HTML tags and DOM structure as first-class signals. Other architectures, notably WebFormer~\cite{WebFormer}, create complex attention patterns between HTML and text tokens, thereby training the model to capture HTML structure. However, such models are restricted by token windows (typically 512 tokens), which limits their capacity to encode entire web pages (see Section~\ref{chap:long_context}). Contemporary state-of-the-art text encoders such as Jina’s jina-embeddings-v3~\cite{jina} and OpenAI’s text-embeddings-3~\cite{openai_embed} provide strong semantic representations, yet their operational costs make inference over large, continuously evolving web corpora expensive. Large language models (LLMs) have also demonstrated promising results on tasks including HTML understanding \cite{UnderstandingHTMLWithLLM}. Nevertheless, their scale makes them impractical for large-scale deployment.

To the best of our knowledge, no existing model explicitly targets the problem of HTML document embedding while simultaneously emphasizing computational efficiency and the large context lengths required to represent real-world web pages. Furthermore, existing approaches predominantly focus on English, whereas our work primarily targets the Czech language, reflecting the needs of Seznam.cz as a Czech-based search engine. In this work, we present HTML-LM, a model that efficiently compresses web page content into high-quality embeddings suitable for a wide range of classification and regression tasks.

\section{Methodology}

\subsection{Exploratory Analysis and Design Choices}
We conducted a comprehensive exploratory analysis to guide our decisions. First, we developed a suite of benchmarks (see Section~\ref{chap:dws_performance}) to enable systematic comparisons across models. Using these benchmarks, we evaluated several state-of-the-art models without any task-specific fine-tuning.

This evaluation identified Qwen3-Embedding-8B~\cite{Qwen3Embedding} as a strong performer, supporting the feasibility of universal HTML document representations. Further analyses revealed two critical factors for effective HTML processing: (i) a minimum context window of 2048 tokens (see Section~\ref{chap:long_context}), and (ii) explicit preservation of HTML structure (see Section~\ref{chap:html_vs_text}).

\subsection{Data} \label{chap:data}
We trained our model on a large corpus of 100 million HTML documents. The dataset comprised 53\% Czech domains (\texttt{.cz}), 34\% primarily English-language domains (\texttt{.com}, \texttt{.org}, \texttt{.net}), 9\% other European domains (\texttt{.sk}, \texttt{.de}, \texttt{.eu}, \texttt{.pl}, \texttt{.it}, \texttt{.at}), and the remaining 4\% from less common domains. The dataset was sampled from our internal web database, which stores crawled pages along with related metadata, including text length, domain, and document cluster. We used these metadata to systematically design and refine a sampling strategy. After extensive experimentation, our sampling procedure produced a dataset with text lengths uniformly distributed, a limit of 1000 documents per domain, and clusters with high entropy prioritized.

\subsubsection{Pre-processing} We adopted an aggressive preprocessing strategy similar to~\cite{Webformer2022}. When tokenized, preprocessed HTML typically retains fewer than 5\% of the original tokens while preserving most of the information needed for accurate downstream task processing. The algorithm operates as follows:

\newcommand\textvtt[1]{{\normalfont\fontfamily{cmvtt}\selectfont #1}}

\begin{enumerate}
    \item \textbf{HTML Parsing}: The HTML document is parsed using \textvtt{BeautifulSoup4} with the \texttt{lxml} backend, producing a Document Object Model (DOM) tree.

    \item \textbf{DOM Pruning}: During a depth-first DOM traversal, the tree is processed
as follows:
    \begin{itemize}
        \item \textbf{Removal}: We remove all subtrees without plain-text content as well as all \texttt{<script>}, \texttt{<style>}, and comment nodes.
        
        \item \textbf{Retention}: All textual content is preserved, including tags around it. In addition, the following tags are explicitly retained even when they do not contain text, as they encode relevant structural or semantic information: \texttt{a, address, audio, br, button, col, embed, figure, footer, form, frame, header, hr, iframe, img, input, label, \\ menu, meter, nav, option, output, picture, progress, search, select, td, th, tr, textarea, video, i, svg}.
        
        \item \textbf{Attribute Stripping}: Finally, attributes are removed from each retained node.
    \end{itemize}

    \item \textbf{DOM Compression}: Nodes with exactly one child are replaced by their child, reducing unnecessary hierarchy.

    \item \textbf{Whitespace Normalization}: Multiple consecutive whitespace characters are collapsed into a single space.
\end{enumerate}

\subsubsection{Tokenization}
To improve tokenization efficiency, we created a custom HTML-optimized WordPiece tokenizer \cite{wordpice} with a 57K-token vocabulary and a minimum token frequency of 10, balancing sequence length (benefiting from larger vocabularies) against model size (the embedding matrix size). Using our training corpus, we identified the 100 most frequent HTML tags and assigned dedicated tokens for their opening and closing forms. The number was selected empirically; including more than 100 tags provided no gains, while keeping only the top 50 tags worsened the tokenization performance. The tokenizer was trained from scratch on 5M randomly sampled documents from the model training corpus (see Section~\ref{chap:data}), as larger subsets did not improve training without providing measurable benefits.
    
\subsection{Model}
Our model is based on the ModernBERT~\cite{ModerBERT} architecture, which supports context windows of 8192 tokens and allows for efficient processing of long documents. The proposed HTML-LM model consists of 22 layers and 12 attention heads, employs a hidden size of 768, and contains a total of 154 million parameters. Because we replaced the tokenizer and modified key hyperparameters, the model was trained from scratch for a single complete pass over the training dataset using the Adam optimizer with a learning rate of $2.5 \cdot 10^{-5}$. Training employed a trapezoidal linear learning-rate schedule with warm-up and cooldown phases covering 10\% and 20\% of the total training steps, respectively. The maximum sequence length (during training) was limited to 4096 tokens, and the batch size was set to 32 per GPU, yielding a global batch size of 256 distributed across 8 NVIDIA H100 GPUs.

\subsection{Training Objectives} We trained our model in a multi-task setting with objectives, including Masked Language Modeling \cite{bert_mlm}, the Bag-of-Word Prediction~\cite{BOWLoss}, and contrastive distillation of LLM embeddings~\cite{contrastive_loss_info_nce}.

\subsubsection{Projection Heads}
To make loss functions computable, embeddings from the model’s hidden dimension $D$ are projected into loss-specific target spaces of dimension $T$.

\begin{itemize}
    \item \textit{Language Modeling (LM) and Bag-of-Words (BOW) Heads} map embeddings from $D$ to the full vocabulary space $|V|$. Instead of introducing a new projection layer, we use the transpose of the model’s embedding matrix~\cite{EmbeddingMatrix}. To further improve efficiency, we use the Cut Cross-Entropy (CCE)~\cite{CutCrossEntropy}.
    
    \item \textit{Distillation Head} projects embeddings from $D$ to the teacher space $T$ via
a low-rank factorization (Equation 1). The rank $R$ reduces the number of
additional parameters while approximating the full $D \times T$ projection. This
head is discarded after training.
    
\end{itemize}

\begin{equation} \label{eq:factorized_head}
    W^{D \times T} \approx W^{D \times R} \times W^{R \times T} \quad \text{where} \quad R < D < T
\end{equation}

\subsubsection{Masked Language Modeling (MLM)} Following standard practice, we adopt MLM as a training objective. By masking both textual and HTML tokens, the model learns to capture both semantic content and structural information from the DOM, while reinforcing token-level representations.

\subsubsection{Bag-of-Words Prediction (BOW)} BOW operates on masked inputs like MLM, but instead of reconstructing each token from its local context, it recovers all tokens from the [CLS] representation, encouraging a global, document-level understanding.

\subsubsection{LLM Distillation}
To incorporate knowledge from substantially larger teacher models, we employ \emph{Teacher-Space Weighted Contrastive Distillation}, using Qwen3-Embedding-8B~\cite{Qwen3Embedding} and SeLLMa 8B (internal LLM model based on Llama~3.1\footnote{\url{https://huggingface.co/meta-llama/Llama-3.1-8B}}~\cite{LLama3} and fine-tuned for the Czech language)~\cite{sellma_1}~\cite{sellma_2} as teachers. Unlike standard in-batch contrastive losses such as InfoNCE~\cite{contrastive_loss_info_nce} and SimCLR~\cite{simclr}, which treat all negatives equally, our method weights each negative pair based on teacher guidance. Negatives considered similar by the teacher are penalized less, allowing closer representations. Our implementation builds on SoftCSE~\cite{SoftCSE} but introduces a different teacher guidance scheme. Formally, for a batch $B$, we define the distillation loss in Equation~\ref{eq:distil_loss}.

\begin{equation} \label{eq:distil_loss}
    \mathcal{L}_{\text{distil}} = 
        \mathbb{E}_{i\in B} \Bigg[
        -\log 
        \frac
            {\exp\left(s_{i,i} / \tau\right)}
            {\exp\left(s_{i,i} / \tau\right) + \sum_{j \in B, j \neq i}  w_{i,j} \, \exp\left(s_{i,j} / \tau\right)}
        \Bigg]
\end{equation}

Here, $s_{i,j}$ denotes the cosine similarity between the student's embedding of document $i$ and the teacher's embedding of document~$j$. The teacher guidance weight $w_{i,j} = \frac{1 - \tilde{s}_{i,j}}{2} \in [0,1]$ is based on $\tilde{s}_{i,j}$, which represents the cosine similarity between the teacher's representation of documents $i$ and $j$. Finally, the temperature $\tau$ controls the sharpness of the distribution. We found $\tau=0.1$ to perform the best.

\subsubsection{Aggregation} When using multiple loss functions during training, these losses must be unified into a single objective $\mathcal{L}$.

\begin{equation} \label{eq:loss_comb}
    \mathcal{L} = \sum \lambda_i \cdot \alpha_i  \cdot \frac{\mathcal{L}_i}{\mathcal{L}_i^{initial}} \quad \alpha_i \sim \mathcal{U}(0, 1)
\end{equation}

Because individual losses differ in scale, we first estimate the magnitude of each loss from the initial batch, $\mathcal{L}_i^{\text{initial}}$, and then normalize it so that all losses affect the overall objective to a similar extent. According to Equation~\ref{eq:loss_comb}, the total loss $\mathcal{L}$ is formulated as a weighted sum of the individual normalized loss terms. Additionally, each loss term is scaled by a fixed baseline weight $\lambda_i$ ($\lambda_{\text{MLM}} = 0.1, \lambda_{\text{BOW}} = 1, \lambda_{\text{distil}} = 1$), which specifies the relative importance of each objective, and a stochastic coefficient $\alpha_i$ drawn uniformly from $\mathcal{U}(0, 1)$ at every step. This strategy helps the model prioritize different training aspects at each step, thereby improving generalization. \cite{random_weighting}

\section{Evaluation}

\subsection{Tokenizer Performance} \label{chap:tokenizer}

\begin{table}
    \centering
    \begin{tabular}{l|c|c|c|c|c}
        \textbf{Tokenizer} & \textbf{Size} & $\mathbf{\leq 512}$ & $\mathbf{\leq 4096}$ & $\mathbf{\leq 8192}$ & \textbf{Util.}  \\
        \hline
        MarkupLM$^{**}$                 & 50K & 25.8 & 88.7 & 96.3 & \textbf{99.5} \\
        tiktoken cl100k\_base$^*$       & 100K & 10.8 & 80.6 & 93.5 & 84.1 \\
        ModernBERT                      & 50K & 10.0 & 79.3 & 92.9 & 97.3 \\
        jina-embeddings-v3              & 250K & 9.8 & 80.7 & 93.6 & 77.7 \\
        Qwen3-Embedding-8B              & 151K & 10.4 & 80.2 & 93.4 & 78.6 \\
        SeLLMa 8B                       & 128K & \textbf{29.2} & \textbf{92.2} & \textbf{97.5 }& 86.2 \\
        RetroMAE \cite{retromae}                 & 57K & 7.9 & 77.2 & 92.1 & 97.0 \\
        \textbf{HTML-LM (ours)}          & 57K & 10.3 & 83.7 & 94.8 & 98.9 \\
        \hline
    \end{tabular}
    \caption{Tokenizer performance, measured as the percentage of documents that, when tokenized, fully fit within context windows of varying lengths ($\leq N$). The \emph{Util.} column represents vocabulary utilization, while the \emph{Size} column indicates the vocabulary size.
    \\ $^*$ Used by the OpenAI text-embeddings-3.
    \\ $^{**}$ Does not tokenize HTML tags.
    }
    \label{tab:tokenizer}
\end{table}

To evaluate the proposed tokenizer, we measured the percentage of documents that fit entirely within a given context window after tokenization. This evaluation was performed on a random sample of 100K documents from the model training corpus (see Section~\ref{chap:data}). As shown in Table~\ref{tab:tokenizer}, our tokenizer demonstrates robust performance across all context windows, outperforming all competitors except MarkupLM and SeLLMa 8B. While MarkupLM delivers slightly better performance, it depends on extracting structured information—specifically the XPath of each HTML node—from the DOM, which is computationally expensive. Similarly, SeLLMa’s superior performance is likely driven by its large 128K vocabulary. Consequently, neither is suitable for our specific setting. Overall, among usable tokenizers, our approach achieves the best coverage across all evaluated context window sizes.

\subsection{Evaluation Methodology} \label{chap:dws_performance}
The model performance is assessed across multiple downstream applications, with an emphasis on documents from the Czech domain. During evaluation, the model under investigation remains completely frozen, and only a lightweight, task-specific MLP head is trained on its embeddings. Depending on the task, each head introduces approximately 0.3–1M additional parameters. As presented in Table~\ref{tab:downstream}, our model, although significantly smaller than its main competitors, outperforms all other models. 

\subsubsection{Downstream Applications}
\begin{enumerate}
    \item \textbf{Article Type} (multi-class classification, metric: F1 macro)
        \begin{itemize}
            \item Evaluates the model’s ability to classify article type.
            \item Classes: \textit{Not an article, Tabloid, Journalism, Hobby, Sport, News – local, News – domestic \& international.}
        \end{itemize}
    \item \textbf{Curlie} (multi-class classification, metric: Accuracy)
        \begin{itemize}
            \item Evaluates the model’s ability to classify Czech webpages according to
    the top-level Curlie categories\footnote{outsourced from \url{https://curlie.org/cs}}.
            \item Categories: \textit{Arts, Business, Computers, Health, Home, News, Science,
    Sports, Shopping, Kids and Teens.}
        \end{itemize}
    \item \textbf{Porn} (multi-class classification, metric: F1 macro)
        \begin{itemize}
            \item Evaluates the model’s ability to classify explicit content.
            \item Classes: \textit{Safe, Adult, Porn}.
        \end{itemize}
    \item \textbf{Product} (multi-class classification, metric: F1 macro)
        \begin{itemize}
            \item Evaluates the model’s ability to identify product-related pages.
            \item Classes: \textit{E-shop product list, E-shop product detail, Other}.
        \end{itemize}
    \item \textbf{Web Spam} (regression, metric: RMSE)
        \begin{itemize}
            \item Measures how well the model can predict the amount of spam in the
    page.
            \item Output: continuous score representing the spam amount.
        \end{itemize}
\end{enumerate}

\begin{table*}[!t]
    \centering
    \resizebox{\textwidth}{!}{
        \begin{tabular}{l|c|c|c|c|c|c|c}
        \hline
            \textbf{Model} 
            & \textbf{Params} 
            & \textbf{Article Type} 
            & \textbf{Curlie} 
            & \textbf{Porn} 
            & \textbf{Product}
            & \textbf{Web Spam}
            & \textbf{Aggregated} \\
            
            metric &
            &
            F1 macro $\uparrow$ &
            Accuracy $\uparrow$ &
            F1 macro $\uparrow$ &
            F1 macro $\uparrow$ &
            RMSE $\downarrow$ &
            NMM  $\uparrow$\\
        \hline
        random                              & 0     & 0.0691 & 0.0328 & 0.3322 & 0.1914 & 0.5619 & 0.0000 \\
        MarkupLM base                       & 135M  & 0.6853 & 0.2198 & 0.5684 & 0.7943 & 0.3120 & 0.4799 \\
        ModernBERT base                     & 149M  & 0.7085 & 0.3177 & 0.4864 & 0.8128 & 0.3166 & 0.4835 \\
        OpenAI text-embeddings-3 small      & ?     & 0.7650 & 0.7406 & 0.6348 & 0.8513 & 0.2678 & 0.6544 \\
        jina-embeddings-v3 base             & 570M  & 0.7856 & 0.7346 & 0.6250 & 0.8753 & 0.3070 & 0.6466 \\
        Qwen3-Embedding                     & 8B  & 0.7732 & 0.7536 & 0.6442 & 0.8937 & 0.3102 & 0.6571 \\
        SeLLMa 8B                           & 8B  & 0.7568 & 0.7406 & 0.5592 & 0.7933 & 0.2827 & 0.6103 \\
        \textbf{HTML-LM base}               & 154M  &\textbf{0.7978} & \textbf{0.7577} & \textbf{0.6521} & \textbf{0.9188} & \textbf{0.2306} & \textbf{0.7001} \\
        \hline
        \end{tabular}
    }
    \caption{Comparison of model performance across all downstream tasks. All models were evaluated using HTML-preserving inputs and an 8192-token context window, except for MarkupLM, which supports only 512 tokens, and the random model, which receives no input. The random model’s scores were derived from heads trained on random embeddings.}
    \label{tab:downstream}
\end{table*}

\subsubsection{Normalized Metric Mean (NMM)}
We use the \textbf{Normalized Metric Mean (NMM)} to evaluate performance across multiple downstream applications. This metric represents the average improvement of a model over a baseline $R$. For each task $t$, we calculate a contribution $C_t$ based on the model performance $M_t$ relative to the baseline $R_t$, where $C_t = \frac{M_t - R_t}{1 - R_t}$ if $M_t$ is maximized, and~$C_t = 1 - \frac{M_t}{R_t}$ if~$M_t$ is minimized. These contributions are then averaged across all~$T$ tasks, yielding a single aggregated metric that facilitates easy comparison of multiple models. Using a random model as the baseline, the NMM quantifies the extent to which our approach outperforms chance. 

\subsection{Long Context} \label{chap:long_context}
Following the methodology outlined in Section~\ref{chap:dws_performance}, we examined the context length required for effective HTML document processing by measuring performance on inputs truncated to 512, 2048, 4096, and 8192 tokens. As shown in Table~\ref{tab:long_context}, Qwen3-Embedding-8B demonstrates consistent performance improvements up to 4096 tokens, after which performance stabilizes. Our HTML-LM model shows a similar trend, delivering strong performance at 2048 tokens, with continued improvement as the context length increases up to the model’s maximum of 8192 tokens.

\begin{table}
    \centering
    \begin{tabular}{l|c|c|c|c}
        \textbf{Model} & $\mathbf{\leq 512}$ & $\mathbf{\leq 2048}$ & $\mathbf{\leq 4096}$ & $\mathbf{\leq 8192}$  \\
        metric & NMM  $\uparrow$ & NMM  $\uparrow$ & NMM  $\uparrow$ & NMM  $\uparrow$ \\
        \hline
        Qwen3-Embedding (8B)            & 0.6327 & 0.6486 & \textbf{0.6581} & 0.6571 \\
        \textbf{HTML-LM base (154M)}     & 0.6807 & 0.6952 & 0.6954 & \textbf{0.7001} \\
        \hline
    \end{tabular}
    \caption{Model performance measured on HTML-preserving inputs truncated to varying context lengths ($\leq N$).}
    \label{tab:long_context}
\end{table}

\subsection{HTML vs. Plain Text} \label{chap:html_vs_text}
Following the methodology outlined in Section~\ref{chap:dws_performance}, we assessed the effect of preserving HTML markup on model performance. We compared models on content presented in two forms: with HTML tags preserved and with HTML tags removed (leaving only plain text). As shown in Table~\ref{tab:html_vs_text}, preserving HTML tags improves performance for the Qwen3-Embedding-8B model, even though the model was not explicitly trained on HTML. Motivated by this finding, we trained the proposed HTML-LM model on HTML-preserving inputs and, for comparison, trained an identical model on plain-text inputs. This direct comparison—using the same architecture and training setup, differing only in input format—demonstrates that preserving HTML improves performance. In the HTML-LM setup, preserving HTML improves NMM by 0.0064, a gain roughly comparable to increasing the model size from 75M to 154M parameters.

\begin{table}
    \centering
    \begin{tabular}{l|c|c}
        \textbf{Model} & \textbf{HTML} & \textbf{Text}  \\
        metric & NMM  $\uparrow$ & NMM  $\uparrow$ \\
        \hline
        Qwen3-Embedding (8B)                & \textbf{0.6571} & 0.6504      \\
        \textbf{HTML-LM base (154M)}        & \textbf{0.7001} & 0.6937      \\
        \hline
    \end{tabular}
    \caption{Comparison of model performance on inputs with HTML preserved (HTML) versus inputs with HTML removed (Text). Models were evaluated using an 8192-token context.}
    \label{tab:html_vs_text}
\end{table}

\section{Conclusion}
We presented HTML-LM, a compact, HTML-aware foundation model designed to generate versatile, high-quality representations of web documents, with a focus on the Czech Internet domain. By leveraging a HTML-informed training, an aggressive preprocessing pipeline, and a HTML-optimized tokenizer, HTML-LM produces generalizable embeddings while remaining computationally efficient. Despite having only 154 million parameters, it achieves state-of-the-art results across multiple Czech classification and regression benchmarks, outperforming both larger embedding models and small-sized LLMs. HTML-LM has also been deployed at scale in production, processing thousands of documents per second, demonstrating strong performance and a real-world industrial impact.

\begin{credits}
\subsubsection{\ackname}
This work was carried out as part of the project HTML-LM, funded by Seznam.cz. This preprint has not undergone peer review (when applicable) or any post-submission improvements or corrections. The Version of Record of this contribution is published in \textbf{Text, Speech, and Dialogue (TSD 2026)}, and is available online at \url{https://doi.org/10.1007/978-3-032-37249-9_12}.

\subsubsection{\discintname}
All authors are employees of Seznam.cz. The study was conducted within the HTML-LM project, and its outcomes are used in the company’s production systems.
\end{credits}

\bibliographystyle{splncs04}
\bibliography{tsd1396a}

@inproceedings{Webformer2022,
  title={Webformer: Pre-training with web pages for information retrieval},
  author={Guo, Yu and Ma, Zhengyi and Mao, Jiaxin and Qian, Hongjin and Zhang, Xinyu and Jiang, Hao and Cao, Zhao and Dou, Zhicheng},
  booktitle={Proceedings of the 45th International ACM SIGIR Conference on Research and Development in Information Retrieval},
  pages={1502--1512},
  year={2022},
  doi={10.1145/3477495.3532086},
}

@inproceedings{BOWLoss,
  title={Drop your Decoder: Pre-training with Bag-of-Word Prediction for Dense Passage Retrieval.},
  author={Ma, Guangyuan and Wu, Xing and Lin, Zijia and Hu, Songlin},
  booktitle={Proceedings of the 47th International ACM SIGIR Conference on Research and Development in Information Retrieval},
  pages={1818--1827},
  year={2024},
  doi={10.1145/3626772.3657792},
}

@inproceedings{ModerBERT,
  title={Smarter, better, faster, longer: A modern bidirectional encoder for fast, memory efficient, and long context finetuning and inference},
  author={Warner, Benjamin and Chaffin, Antoine and Clavi{\'e}, Benjamin and Weller, Orion and Hallstr{\"o}m, Oskar and Taghadouini, Said and Gallagher, Alexis and Biswas, Raja and Ladhak, Faisal and Aarsen, Tom and others},
  booktitle={Proceedings of the 63rd Annual Meeting of the Association for Computational Linguistics (Volume 1: Long Papers)},
  pages={2526--2547},
  year={2025},
  doi={10.18653/v1/2025.acl-long.127},
}

@inproceedings{SoftCSE,
  title={Not all negatives are equally negative: Soft contrastive learning for unsupervised sentence representations},
  author={Zhuang, Haojie and Emma Zhang, Wei and Yang, Jian and Chen, Weitong and Sheng, Quan Z},
  booktitle={Proceedings of the 33rd ACM International Conference on Information and Knowledge Management},
  pages={3591--3601},
  year={2024},
  doi={10.1145/3627673.3679745},
}

@article{DOMLM,
  title={Dom-lm: Learning generalizable representations for html documents},
  author={Deng, Xiang and Shiralkar, Prashant and Lockard, Colin and Huang, Binxuan and Sun, Huan},
  journal={arXiv preprint arXiv:2201.10608},
  year={2022},
  doi={10.48550/arXiv.2201.10608},
}

@inproceedings{markuplm,
  title={MarkupLM: Pre-training of text and markup language for visually rich document understanding},
  author={Li, Junlong and Xu, Yiheng and Cui, Lei and Wei, Furu},
  booktitle={Proceedings of the 60th Annual Meeting of the Association for Computational Linguistics (Volume 1: Long Papers)},
  pages={6078--6087},
  year={2022},
  doi={10.18653/v1/2022.acl-long.420},
}

@article{jina,
  title={jina-embeddings-v3: Multilingual embeddings with task lora},
  author={Sturua, Saba and Mohr, Isabelle and Akram, Mohammad Kalim and G{\"u}nther, Michael and Wang, Bo and Krimmel, Markus and Wang, Feng and Mastrapas, Georgios and Koukounas, Andreas and Wang, Nan and others},
  journal={arXiv preprint arXiv:2409.10173},
  year={2024},
  doi={10.48550/arXiv.2409.10173},
}

@article{Qwen3Embedding,
  title={Qwen3 Embedding: Advancing Text Embedding and Reranking Through Foundation Models},
  author={Zhang, Yanzhao and Li, Mingxin and Long, Dingkun and Zhang, Xin and Lin, Huan and Yang, Baosong and Xie, Pengjun and Yang, An and Liu, Dayiheng and Lin, Junyang and others},
  journal={arXiv preprint arXiv:2506.05176},
  year={2025},
  doi={10.48550/arXiv.2506.05176},
}

@inproceedings{WebFormer,
  title={WebFormer: The Web-page Transformer for Structure Information Extraction},
  author={Wang, Qifan and Fang, Yi and Ravula, Anirudh and Feng, Fuli and Quan, Xiaojun and Liu, Dongfang},
  booktitle={Proceedings of the ACM Web Conference 2022},
  pages={3124--3133},
  year={2022},
  doi={10.1145/3485447.3512032},
}

@article{CutCrossEntropy,
  title={Cut your losses in large-vocabulary language models},
  author={Wijmans, Erik and Huval, Brody and Hertzberg, Alexander and Koltun, Vladlen and Kr{\"a}henb{\"u}hl, Philipp},
  journal={arXiv preprint arXiv:2411.09009},
  year={2024},
  doi={10.48550/arXiv.2411.09009},
}

@inproceedings{EmbeddingMatrix,
  title={Using the output embedding to improve language models},
  author={Press, Ofir and Wolf, Lior},
  booktitle={Proceedings of the 15th Conference of the European Chapter of the Association for Computational Linguistics: Volume 2, Short Papers},
  pages={157--163},
  year={2017},
  doi={10.48550/arXiv.1608.05859},
}

@article{UnderstandingHTMLWithLLM,
  title={Understanding html with large language models. arxiv 2022},
  author={Gur, I and Nachum, O and Miao, Y and Safdari, M and Huang, A and Chowdhery, A and Narang, S and Fiedel, N and Faust, A},
  journal={arXiv preprint arXiv:2210.03945},
  year={2022},
  doi={10.48550/arXiv.2210.03945},
}

@inproceedings{SearchingBasedOnRelevance,
  title={Searching documents based on relevance and type},
  author={Xu, Jun and Cao, Yunbo and Li, Hang and Craswell, Nick and Huang, Yalou},
  booktitle={European Conference on Information Retrieval},
  pages={629--636},
  year={2007},
  organization={Springer},
  doi={10.1007/978-3-540-71496-5_60},
}

@article{SpamToIR,
  title={From spam filtering to information retrieval and back: seeking conceptual foundations for spam filtering},
  author={Lueg, Christopher P},
  journal={Proceedings of the American society for information science and technology},
  volume={42},
  number={1},
  year={2005},
  publisher={Wiley Online Library},
  doi={10.1002/meet.14504201146},
}

@article{LLama3,
  title={The llama 3 herd of models},
  author={Grattafiori, Aaron and Dubey, Abhimanyu and Jauhri, Abhinav and Pandey, Abhinav and Kadian, Abhishek and Al-Dahle, Ahmad and Letman, Aiesha and Mathur, Akhil and Schelten, Alan and Vaughan, Alex and others},
  journal={arXiv preprint arXiv:2407.21783},
  year={2024},
  doi={10.48550/arXiv.2407.21783},
}

@inproceedings{bert_mlm,
  title={Bert: Pre-training of deep bidirectional transformers for language understanding},
  author={Devlin, Jacob and Chang, Ming-Wei and Lee, Kenton and Toutanova, Kristina},
  booktitle={Proceedings of the 2019 conference of the North American chapter of the association for computational linguistics: human language technologies, volume 1 (long and short papers)},
  pages={4171--4186},
  year={2019},
  doi={10.18653/v1/N19-1423},
}

@article{contrastive_loss_info_nce,
  title={Representation learning with contrastive predictive coding},
  author={Oord, Aaron van den and Li, Yazhe and Vinyals, Oriol},
  journal={arXiv preprint arXiv:1807.03748},
  year={2018},
  doi={10.48550/arXiv.1807.03748},
}

@article{random_weighting,
  title={Reasonable effectiveness of random weighting: A litmus test for multi-task learning},
  author={Lin, Baijiong and Ye, Feiyang and Zhang, Yu and Tsang, Ivor W},
  journal={arXiv preprint arXiv:2111.10603},
  year={2021},
  doi={10.48550/arXiv.2111.10603},
}

@inproceedings{simclr,
  title={A simple framework for contrastive learning of visual representations},
  author={Chen, Ting and Kornblith, Simon and Norouzi, Mohammad and Hinton, Geoffrey},
  booktitle={International conference on machine learning},
  pages={1597--1607},
  year={2020},
  organization={PmLR},
  doi={10.48550/arXiv.2002.05709},
}

@article{wordpice,
  title={Google's neural machine translation system: Bridging the gap between human and machine translation},
  author={Wu, Yonghui and Schuster, Mike and Chen, Zhifeng and Le, Quoc V and Norouzi, Mohammad and Macherey, Wolfgang and Krikun, Maxim and Cao, Yuan and Gao, Qin and Macherey, Klaus and others},
  journal={arXiv preprint arXiv:1609.08144},
  year={2016},
  doi={10.48550/arXiv.1609.08144},
}

@inproceedings{structor,
  title={Learning structural co-occurrences for structured web data extraction in low-resource settings},
  author={Zhang, Zhenyu and Yu, Bowen and Liu, Tingwen and Liu, Tianyun and Wang, Yubin and Guo, Li},
  booktitle={Proceedings of the acm web conference 2023},
  pages={1683--1692},
  year={2023},
  doi={10.1145/3543507.3583387},
}

@misc{openai_embed,
	author = {OpenAI},
	month = {1},
	title = {{New embedding models and API updates}},
	year = {2024},
	url = {https://openai.com/index/new-embedding-models-and-api-updates/?utm_source=chatgpt.com},
}

@misc{sellma_1,
  title        = {Diana Hlaváčová: SeLLMa aneb Jak v Seznamu krotíme dravé jazykové modely?},
  author       = {Seznam.cz},
  year         = 2024,
  howpublished = {\url{https://blog.seznam.cz/2024/10/diana-hlavacova-sellma-aneb-jak-v-seznamu-krotime-drave-jazykove-modely/}},
  note         = {Accessed: 2026-02-09}
}

@misc{sellma_2,
  title        = {Peter Pekarovič and Martin Kirschner: Seznam AI. Technologie, která není jen chytrá, ale hlavně užitečná},
  author       = {Seznam.cz},
  year         = 2025,
  howpublished = {\url{https://blog.seznam.cz/2025/10/peter-pekarovic-martin-kirschner-seznam-ai-technologie-ktera-neni-jen-chytra-ale-hlavne-uzitecna/}},
  note         = {Accessed: 2026-02-09}
}

@inproceedings{retromae,
  title={Some like it small: Czech semantic embedding models for industry applications},
  author={Bedn{\'a}{\v{r}}, Ji{\v{r}}{\'\i} and N{\'a}plava, Jakub and Baran{\v{c}}{\'\i}kov{\'a}, Petra and Lisick{\`y}, Ond{\v{r}}ej},
  booktitle={Proceedings of the AAAI Conference on Artificial Intelligence},
  volume={38},
  pages={22734--22742},
  year={2024},
  doi={10.1609/aaai.v38i21.30307},
}

\end{document}